\documentclass[10pt,twocolumn,letterpaper]{article}

\usepackage[final]{cvpr}       
\usepackage{booktabs}
\usepackage{todonotes}
\usepackage{amsmath, amssymb}

\PassOptionsToPackage{prologue,dvipsnames}{xcolor}
\usepackage[dvipsnames]{xcolor}

\definecolor{cvprblue}{rgb}{0.21,0.49,0.74}
\usepackage[pagebackref,breaklinks,colorlinks,citecolor=cvprblue]{hyperref}
\usepackage{float}
\def\paperID{15384} 
\def\confName{CVPR}
\def\confYear{2024}
\begin{document}
\title{3DAxiesPrompts: Unleashing the 3D Spatial Task Capabilities of GPT-4V}
\author{
    Dingning Liu$^{1,2}$\\
    \tt \small{dutldn@mail.dlut.edu.cn}
    \and Xiaomeng Dong$^{3}$
    \and Renrui Zhang$^{1,4}$
    \and Xu Luo$^{5}$
    \and Peng Gao$^{1}$
    \and Xiaoshui Huang$^{1,}$\footnotemark[1]\\
    \tt \small{huangxiaoshui@pjlab.org.cn}
    \and Yongshun Gong$^{6}$
    \and Zhihui Wang$^{2,}$\footnotemark[1]
    \and $^{1}${Shanghai Artificial Intelligence Laboratory}
    \and $^{2}${Dalian University of Technology}
    \and $^{3}${Northeast Normal University}
    \and $^{4}${The Chinese University of Hong Kong}
    \and $^{5}${University of Electronic Science and Technology of China}
    \and $^{6}${Shandong University}
}

\maketitle
\renewcommand{\thefootnote}{\fnsymbol{footnote}}
\footnotetext[1]{Corresponding author}
\begin{abstract}
In this work, we present a new visual prompting method called 3DAxiesPrompts (3DAP) to unleash the capabilities of GPT-4V in performing 3D spatial tasks. Our investigation reveals that while GPT-4V exhibits proficiency in discerning the position and interrelations of 2D entities through current visual prompting techniques, its abilities in handling 3D spatial tasks have yet to be explored. In our approach, we create a 3D coordinate system tailored to 3D imagery, complete with annotated scale information. By presenting images infused with the 3DAP visual prompt as inputs, we empower GPT-4V to ascertain the spatial positioning information of the given 3D target image with a high degree of precision. Through experiments, We identified three tasks that could be stably completed using the 3DAP method, namely, 2D to 3D Point Reconstruction, 2D to 3D point matching, and 3D Object Detection. We perform experiments on our proposed dataset 3DAP-Data, the results from these experiments validate the efficacy of 3DAP-enhanced GPT-4V inputs, marking a significant stride in 3D spatial task execution.
\end{abstract}
\section{Introduction}
\label{sec:intro}

The recent advent of GPT \cite{chatgpt} has spurred a multitude of innovations across both academic circles and industrial sectors \cite{liu2023summary,zhou_chatgpt_2023,lund2023chatting,liang2023holistic}. The beginning of GPT can only support the interactive question-and-answer with users at the text level \cite{radford2018improving,radford2019language,brown2020language,liu2023pre}. However, the human experience of perception is inherently multimodal, encompassing both visual and language elements, among others \cite{baltruvsaitis2018multimodal}. With the continuous advancement of technology, GPT has undergone a remarkable transformation into a versatile multimodal instrument \cite{openai2023gpt4TR,openai2023gpt4SC,bubeck2023sparks,wu2023visual,zhu2023minigpt}. This evolution has expanded its capabilities beyond mere textual analysis, enabling it to comprehend and dissect a broader spectrum of sensory data, including the intricacies of visual imagery. Recently, the advent of GPT-4v (Vision) has garnered considerable interest. The capabilities of GPT-4v include the processing of image inputs, facilitating an understanding of visual data, and the generation of the corresponding text. Some studies \cite{yang2023set,shtedritski2023does} have shown that GPT-4v has rich semantic information and 2D spatial information understanding ability, and can accurately recognize 2D objects through visual prompts.
\begin{figure}
  \centering
    \includegraphics[scale=0.55]
    {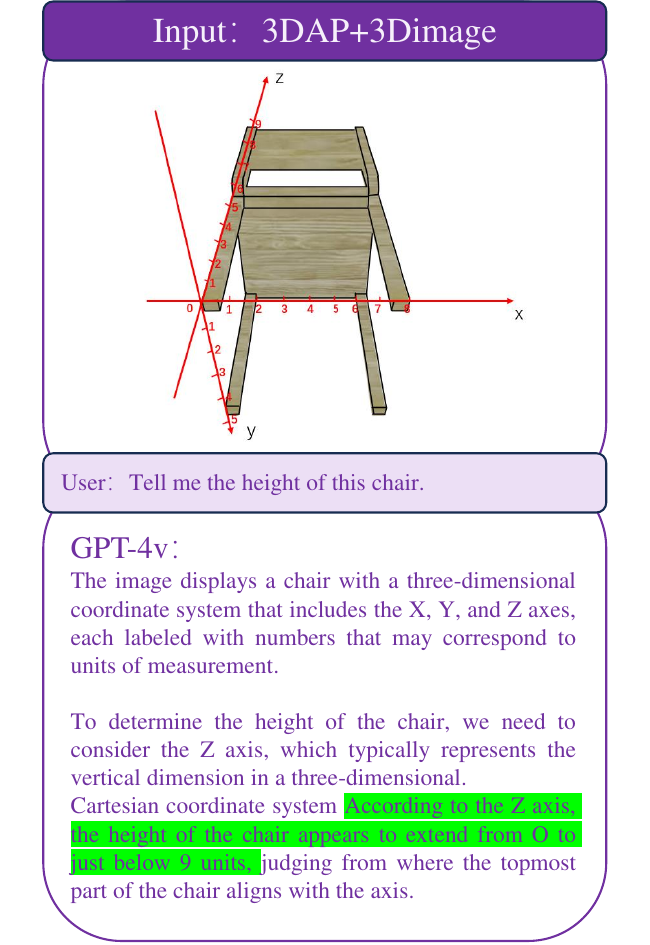}
   \caption{Using the 3DAP method proposed in this paper, GPT-4v accurately answers the height of the stool, the direction axis, and the length of the stool leg.}
   \label{fig:Intro-1}
\end{figure}
 The introduction of GPT-4v brings more possibilities for the task of processing visual information, but at the same time, it also needs continuous research and improvement to improve its 3D modal spatial ability. 3D provides additional depth dimension based on 2D, bringing spatial information significantly closer to the reality we perceive. The depth of information is critical in applications such as autonomous driving, healthcare, navigational mapping, and virtual reality, where three-dimensional data plays a pivotal role in enhancing functionality and realism. Our experiments have led to the finding that GPT-4v falls short in capturing accurate descriptive information from direct three-dimensional image inputs. In response to this challenge, we have introduced a prompting methodology, named 3DAxiesPrompts (3DAP), specifically, by marking the three-dimensional coordinate system and scale information on the input image, meticulously crafted to unleash GPT-4v's potential in comprehending the complexities of three-dimensional environments.

Visual prompts have been employed across an array of visual tasks. In the current research, objects in images are semantically segmented, with each segmented unit assigned a numerical identifier and labeled accordingly \cite{yang2023set}. Visual prompts are also overlaid onto the input image, highlighting crucial areas with red circles \cite{shtedritski2023does}. GPT-4v is adept at locating and mapping the relational dynamics of 2D images guided by these visual prompts, yet its grasp of 3D imagery remains imprecise. The adaptability required of a multimodal model dictates that it must efficiently process various combinations of input modalities. GPT-4v is compatible with three types of input: plain text, single image-text pair, and alternating image-text sequences\cite{openai2023gpt4SC}.

The research motivation of this paper is to stimulate GPT-4v's 3D spatial understanding ability. Concerning the above visual prompts and the input forms supported by GPT-4v, we propose a new visual prompt 3DAxiesPrompts. Specifically, we provided the three-dimensional coordinate system (XYZ-Axies) of input multi-view images
with the scale information according to the unit. The three-dimensional coordinate system with the scale information acts as a spatial mathematical framework, allowing for the description and pinpointing of object positions within the real world. By mapping out these coordinates, we translate the complexities of the real world into quantifiable and computable figures, thereby enabling GPT-4v to analyze objects and their interrelations within the constructed physical space. To demonstrate the efficacy of 3DAxiesPrompts in enhancing GPT-4v's capabilities in 3D space tasks, we present in Figure \ref{fig:Intro-1} how the system can more precisely identify the corresponding axis orientation and coordinate positions of objects (evidenced by the accurate recognition of a stool's leg height).

To summarise, we make the following main contributions:
\paragraph{A new visual prompt method.}
We present a new visual prompt, 3DAxiesPrompts (3DAP), for GPT-4v by marking the 3D coordinate system and scale information of the input picture. Empirical evidence from our experiments substantiates that 3DAP maximizes the potential of GPT-4v’s three-dimensional comprehension capabilities.


\paragraph{Three tasks for experiments.}
 We have determined three tasks through experiments that can be stably achieved using the 3DAP method. Moreover, the aforementioned experiments have proven the practicality and effectiveness of the labeling method we proposed.

\paragraph{A 3D visual prompting dataset.}
 We constructed a small data set called 3DAP-Data based on 3DAP visual prompting. This dataset is meticulously curated based on the principles of 3DAP visual prompting, focusing primarily on objects that feature distinct vertical lines and plane relationships.3DAP-Data aims to enhance and evaluate its ability to understand and interpret three-dimensional spatial relationships.



\section{Related Work}
\label{sec:related}

\paragraph{Adapting Large Models to 3D.}
Driven by the increasing data scale and computation resources, large pre-trained models in both language~\cite{devlin2018bert,radford2019language} and 2D vision~\cite{radford2021learning,he2022masked} have attained significant progress. To fully unleash their generalization capabilities, many efforts have been made to adapt large pre-trained models into the 3D domain. As prior works, Image2Point~\cite{xu2021image2point} directly inflates the pre-trained 2D convolution kernels~\cite{he2016deep} for point cloud processing, and PointCLIP series~\cite{zhang2022pointclip, zhu2023pointclip} leverages CLIP~\cite{radford2021learning} for zero-shot 3D classification by projecting 3D shapes into multi-view depth images. Follow-up works mainly regard the pre-trained large models as teacher networks, and conduct knowledge distillation to transfer their rich semantics into 3D tasks. ULIP series~\cite{xue2023ulip,xue2023ulip2} and OpenShape~\cite{liu2023openshape} adopt a contrastive learning paradigm to pre-train a 3D encoder guided by CLIP. I2P-MAE~\cite{zhang2023learning}, ACT~\cite{dong2022autoencoders}, and other works~\cite{guo2023joint,qi2023contrast} introduce cross-modal MAE~\cite{he2022masked} frameworks to reconstruct language or 2D knowledge. Very recently, multi-modal large language models (LLMs)~\cite{zhang2023llama,han2023imagebind,zhu2023minigpt,li2023blip} have attained promising performance across a wide range of vision-language tasks. As a preliminary attempt, some concurrent works such as 3D-LLM~\cite{hong20233d}, Point-LLM~\cite{guo2023point}, and others~\cite{wang2023chat,xu2023pointllm,jatavallabhula2023conceptfusion,huang2022frozen}, fine-tune LLaMA~\cite{touvron2023llama} for 3D-space question answering and captioning, which, however, requires expensive cross-modal training and are only applicable for open-source LLMs. In this paper, we propose a more general visual prompting method, 3DAxiesPrompts (3DAP), which first investigates the potential of GPT-4v for spatial geometry understanding.

\paragraph{Visual Prompt Learning.}
Prompt engineering has emerged as a crucial area of research in natural language processing (NLP)~\cite{liu2023pre,jiang2020can,shin2020autoprompt}, which adapts pre-trained language models to new tasks without modifying the inner structures. Inspired by this, early works in 2D vision propose to append learnable tokens either in image space (VP~\cite{bahng2022exploring}) or embedding space (VPT~\cite{jia2022visual}) for visual prompt learning. CoOp~\cite{zhou2022learning} and follow-up works~\cite{zhou2022conditional,du2022learning} instead utilize prompt tokens in language space to adapt CLIP~\cite{radford2021learning} for few-shot image classification.
Some efforts~\cite{bar2022visual,zhang2023personalize,wang2023images} also follow the in-context learning paradigm in NLP to provide image examples as 2D visual prompts for pre-trained models during inference time.
Another branch of work explores more explicit prompt signals directly on the input images in a training-free manner, such as colors~\cite{yao2021cpt} and circles~\cite{shtedritski2023does}. 
In the 3D domain, P2P~\cite{wang2022p2p} and Point-PEFT~\cite{tang2023point} explore geometry-aware prompting approaches to achieve parameter-efficient 2D-to-3D learning. 
Recently, with the popularity of multi-modal LLMs, SoM~\cite{yang2023set} introduces a straightforward prompting technique to enhance the visual understanding capacity of GPT-4v, by simply overlaying spatial and speakable marks on the images. Motivated by this, our work aims to fully unleash the geometry perception ability of GPT-4v in 3D space, and proposes a specific 3D visual prompting approach, termed 3DAP, which effectively boosts GPT-4v for favorable spatial reasoning over several 3D tasks.

\paragraph{}

\section{Method}
\label{sec:method}
Our research goal is to propose a visual prompt method based on input images, which can maximize the exploitation of GPT-4V's 3D spatial ability. In this chapter, we introduce the definition of prompt in Section 3.1 and then introduce the steps of constructing the 3D visual prompt method and how to apply the prompt method 3DAP to GPT-4v to unleash the 3D spatial task capabilities in Sections 3.2-3.6.
\subsection{Prompt Definition}
GPT-4V supports three input modes: text-only inputs, selection using a single image-text pair, and the selection of interleaved image-text pairs with multiple image inputs.\cite{openai2023gpt4SC}In this article, we explore two types of input involving images. 

The text in an image-text pair can serve as an instruction, similar to describing the image, or as query input for a question in a visual question-answering context. We define the input image as $I\in R^{X*W*3}$, the text describing the image or asking the question is defined as a text description sequence $L_{in}$ of length $l_i$, and our image visual prompt method is expressed as $AP(I_0, I_1,..., I_n)$, the input image text pair is defined as$ \{AP(I_0, I_1,..., I_n)|L_{in}\}$, generating the response text sequence $L_{out}$ of length $l_o$.

\begin{figure*}
  \centering
    \includegraphics[scale=0.32]
    {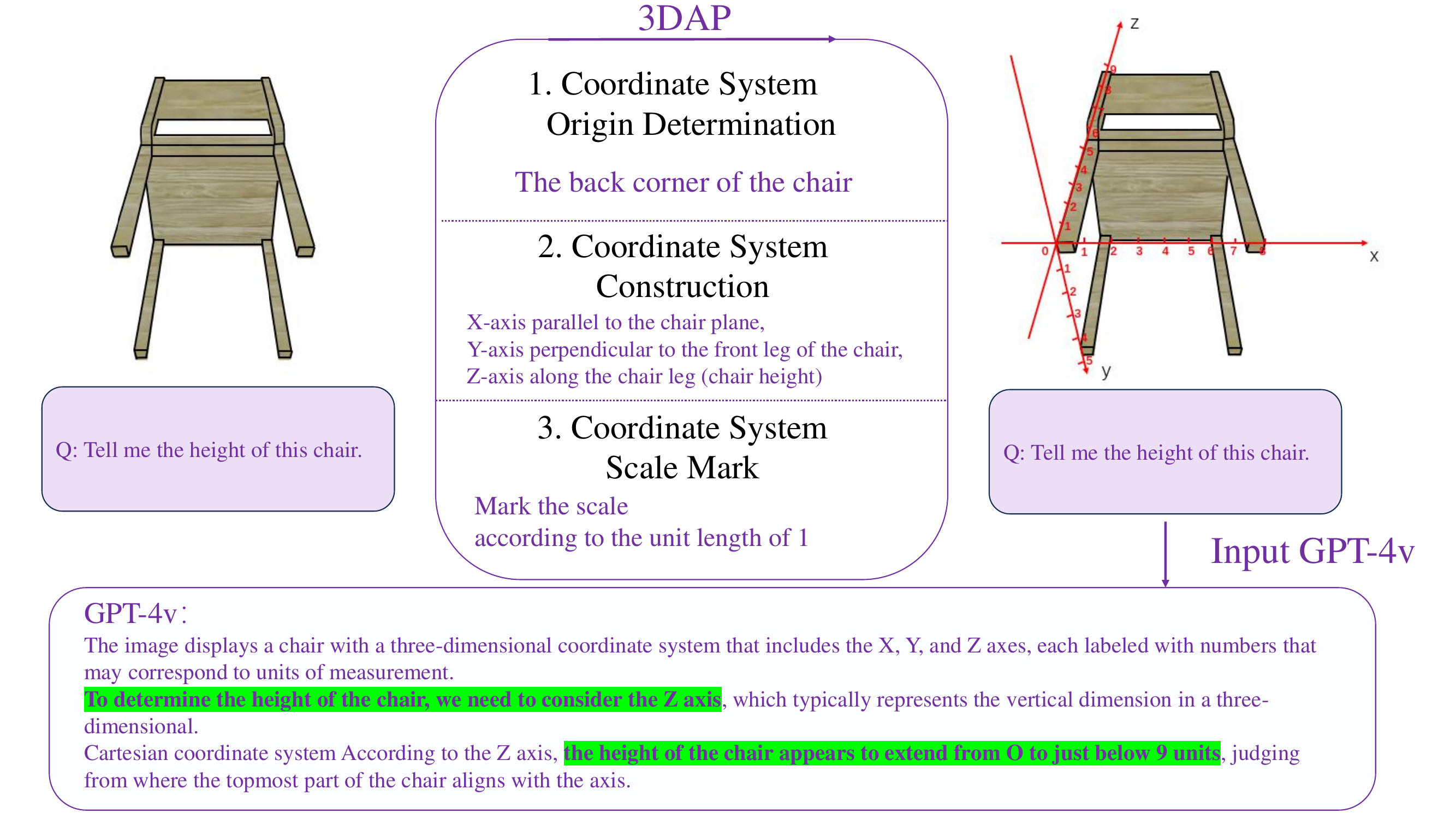}
   \caption{3DAP Example diagram of specific steps in object labeling}
   \label{fig:method-1}
\end{figure*}

\subsection{3D Coordinate System}
The 3D coordinate system, often referred to as the 3D Cartesian coordinate system, extends the two-dimensional Cartesian framework by incorporating an additional dimension that accounts for depth, in addition to the existing horizontal and vertical axes. This system enhances spatial representation by enabling the plotting of points along the front-and-back axis, thus providing a comprehensive model for navigating and visualizing objects within a three-dimensional space.

The three-dimensional coordinate system is structured around a central point known as the origin, denoted by $O$, where its three constituent axes—the X-axis, the Y-axis, and the Z-axis—converge at zero value. This intersection forms the foundational reference point from which the position of any point in the spatial framework is determined, enabling precise localization within the three-dimensional matrix.\cite{soechting1992moving} The three coordinate axes intersect each other perpendicularly. The orthogonal intersection of the three coordinate axes forms a cornerstone for clearly describing and pinpointing an object's location in three-dimensional space.  This enables us to represent and comprehend the object's position, shape, and orientation with precision, providing a foundational system for spatial analysis and visualization.
In a three-dimensional coordinate system,  every point is uniquely defined by a triplet of coordinates $(x, y, z)$, which precisely locates its position within the space. The three-dimensional coordinate system is indispensable in fields such as computer graphics, engineering design, medical imaging, and other advanced scientific and technological domains due to its capacity for precise and clear modeling.
\subsection{Coordinate System Origin Determination}
The origin is a unique fixed point in the three-dimensional coordinate system and is the reference point in the whole coordinate system. This point serves as the commencement of each axis and acts as a reference for locating all other points within the system. The determination of the origin is crucial, as it is directly connected to the relative positioning of other points within the coordinate system. Establishing the origin accurately is fundamental to the system's overall functionality and precision.

In this article, we describe our approach to choosing the origin, which typically relies on the entity's geometric location and directional indicators that are under observation. Many objects, like chairs, beds, tables, and other everyday items, inherently possess fixed orientations. These entities frequently exhibit prominent vertical lines and planes, making their directionality more distinct in everyday life scenes. The position of the origin in the coordinate system is ascertained by establishing the directional axis of our observation.

\subsection{Coordinate System Construction}
In a standard three-dimensional Cartesian coordinate system, the orientation of the axes is typically designated as follows:

The X-axis serves as the horizontal plane, extending rightward to represent the positive direction.

The Y-axis acts as the vertical plane, ascending to signify the positive direction.

The Z-axis corresponds to the depth plane, stretching forward to denote the positive direction.

\begin{figure}
  \centering
    \includegraphics[scale=0.65]
    {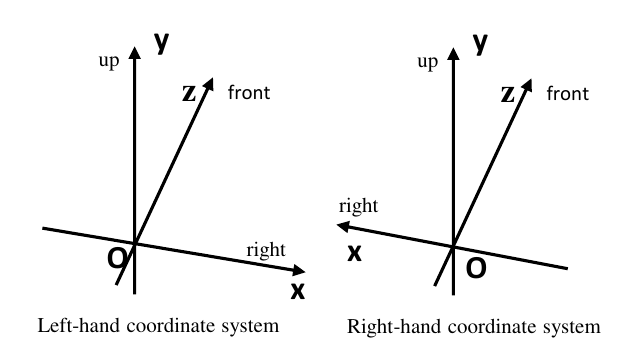}
   \caption{Diagram of left and right-hand coordinate system}
   \label{fig:left-righthand}
\end{figure}

The three axes are mutually perpendicular, creating a three-dimensional space. There are two methods for establishing the orientation of the coordinate axes: the left-hand coordinate system and the right-hand coordinate system. The core concept in both methods is the combination of the three directions of the XYZ axes.\cite{darling1991there}
The left-hand coordinate system usually points the index finger forward, the middle finger to the right extension, and the thumb to the upward extension. The fingers of the right-hand coordinate system point in the same direction, but for the same X and Y axes in the positive direction, the right-hand coordinate system, and the left-hand coordinate system in the opposite direction of the Z axis. Figure \ref{fig:left-righthand} shows the diagram of the left and right-hand coordinate systems. In this paper, we construct the three-dimensional coordinate system according to the left-hand and right-hand coordinate system criteria according to the scene and task requirements.

Each actor in the world scene possesses an absolute coordinate position relative to the world coordinate system, depicting a macroscopic three-dimensional space with constant axis directions. The image represents a projection from the 3D real-world coordinates to 2D plane coordinates. Therefore, recognizing and detecting 3D objects from an image requires reverse transformation into 3D world coordinates, followed by object detection within the world coordinate system. In this paper, the construction of the three-dimensional coordinate system is described above.
\subsection{Coordinate System Scale Mark}
To ensure a precise and uniform approach for quantifying and illustrating spatial values, this paper starts from the origin, marks the scale line with equal spacing along the positive and negative directions of the three coordinate axes, and marks the corresponding unit length value beside the scale line. This facilitates GPT-4V's pinpoint accuracy in identifying and deciphering data point details on three-dimensional entities, thereby enhancing analytical precision.
\subsection{3DAxiesPrompts}
In this section, we introduce the implementation of 3DAP through an example demonstration.

As shown in Figure \ref{fig:method-1}, for an inverted chair, we need to get the height of the chair and the length of the chair leg. We take the back corner of the chair as the origin of the axis, construct the X-axis parallel to the chair plane, construct the Y-axis perpendicular to the front leg of the chair, construct the Z-axis along the chair leg (chair height), and mark the scale information of the axis. Through the three-dimensional coordinate system and scale marking, GPT-4v can accurately construct the spatial features of 3D objects, and identify the axis direction and scale where the features are located.

\section{Experiments}
\label{sec:experiments}
\begin{figure*}
  \centering
    \includegraphics[scale=0.7]
    {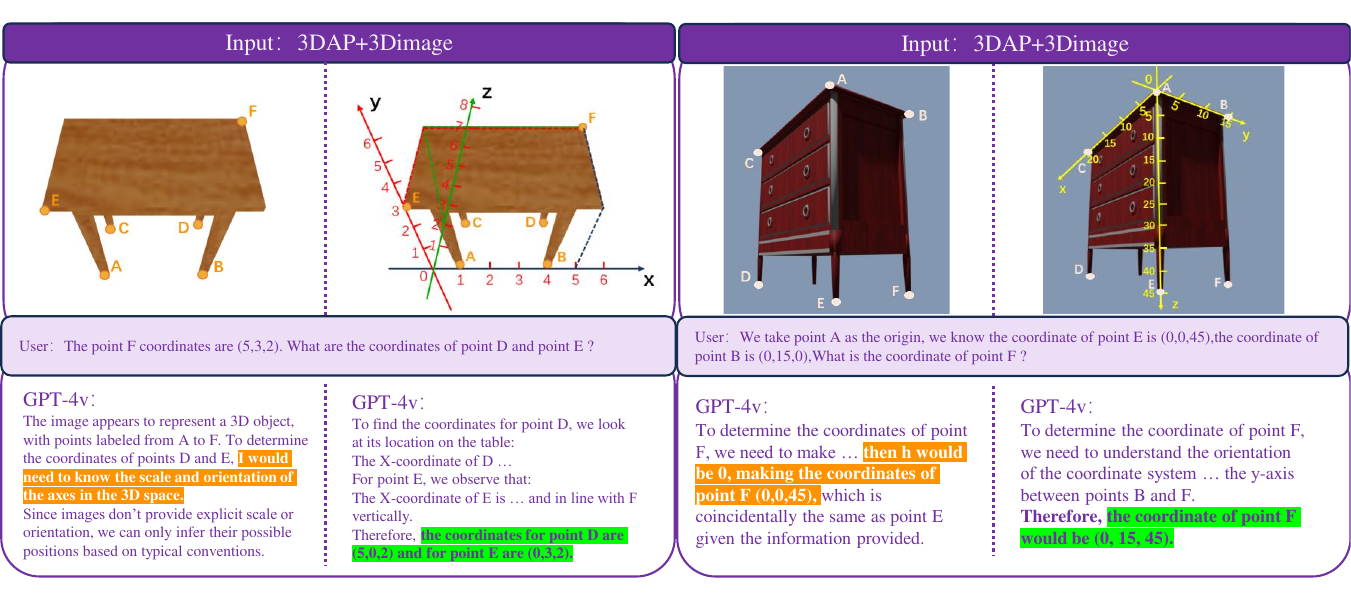}
   \caption{Comparisons of GPT-4V prompting in the task of 2D to 3D Point Reconstruction: (left)we input images that exclusively highlight key point information. Aims to discern the relative positional coordinates of the remaining points. GPT-4V suggests an enhancement: encouraging users to include more comprehensive annotation details, such as directions in the three-dimensional coordinate system and the proportionality of the coordinates, (right) We input 3D images marked with 3DAP, With the coordinate information of a specific point, GPT-4V demonstrates remarkable precision in inferring the relative direction and position coordinates of other points.}
   \label{fig:ex1}
\end{figure*}

\begin{figure*}
  \centering
    \includegraphics[scale=0.6]
    {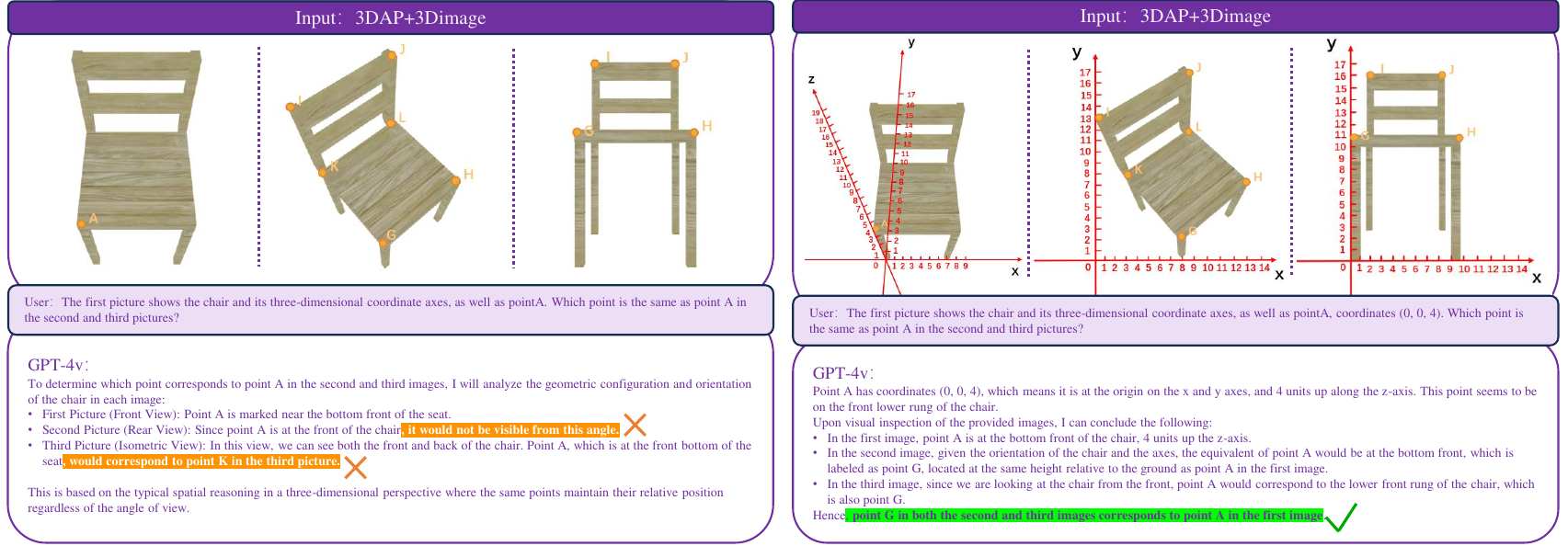}
   \caption{Comparisons of GPT-4V prompting in the task of matching points from 2D to 3D:  (left) We input three images with only marked points, GPT-4V's understanding of spatial information is not accurate, (right) we input one 3D image labeled with 3DAP method and two 2D images, mark the point,  GPT-4V employs its analytical capabilities to ascertain the position of this key point within the 3D coordinate framework, and finds the key point corresponding to the position relationship in the two 2D images.}
   \label{fig:ex2}
\end{figure*}

\subsection{Task Introduction}

This research involved conducting experiments in three key domains: 2D to 3D Point Reconstruction, 2D to 3D Point Matching, and 3D Object Detection. The primary objective was to assess and confirm the stable performance capabilities of the 3DAP system across these varied areas. This section is dedicated to providing a comprehensive overview of these three prevalent 3D tasks.

\textbf{\textit{2D to 3D Point Reconstruction}}
is an advanced image processing technology,  where goal is to extract three-dimensional spatial information from the limited scope of two-dimensional images. \cite{maken20232d,guo2023neural,aharchi2020review,ping2021visual,8850991}The process involves detecting key points or features within these 2D images and then matching these features across various images to establish their three-dimensional spatial correspondence\cite{groueix2018papier,mescheder2019occupancy,wen2019pixel2mesh++,xie2020pix2vox++}. 

The transition from 2D to 3D reconstruction entails identifying feature point information within the 2D image. Subsequently, by analyzing the positions and angles of these feature points across different images, the principles of stereo vision are employed. This involves comparing images captured from diverse perspectives to deduce the object's depth and shape. Such a methodology enables the construction of a precise three-dimensional model that accurately mirrors the spatial form of the object as depicted in the original two-dimensional image. This model then facilitates the inference of the relative positions of the feature points in three-dimensional space, thereby enhancing our understanding and representation of the physical world through digital means.

\begin{figure*}
  \centering
    \includegraphics[scale=0.85]
    {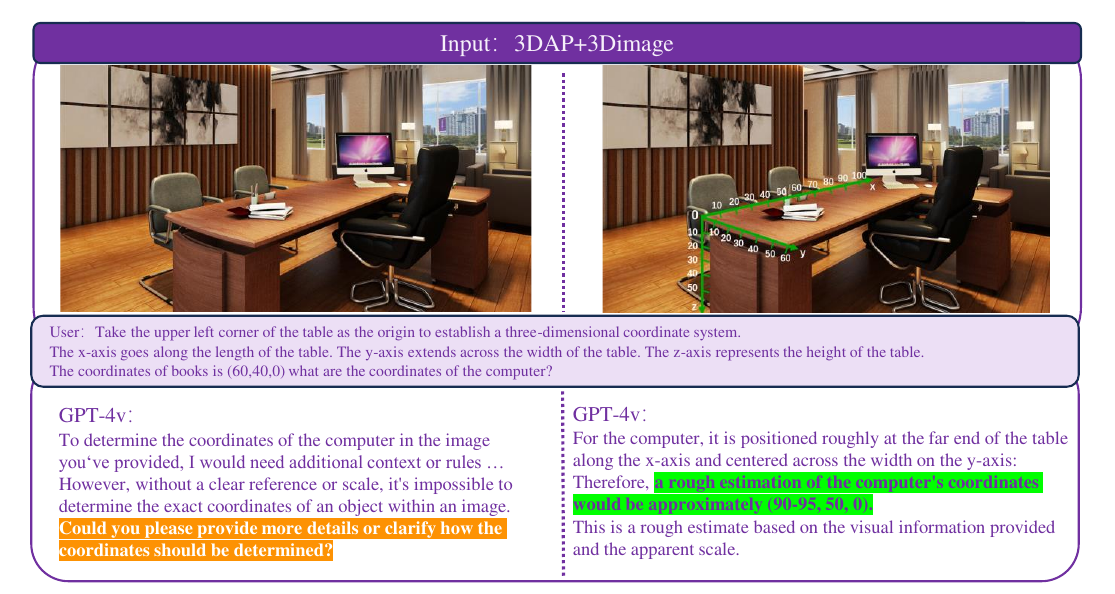}
   \caption{Comparisons of GPT-4V prompting in the task of 3D Object Detection: (left) We input an unmarked image of the original scene presented alongside textual descriptions providing the coordinates of a reference object, GPT-4v prompts the user to input more 3D modeling information, (right) we input the image of the scene is annotated using the 3DAP method, complete with coordinates for the reference object described in the text. GPT-4V can accurately locate object coordinates (ranges) according to the three-dimensional coordinate system and scale size.}
   \label{fig:ex3}
\end{figure*}

\begin{figure*}
  \centering
    \includegraphics[scale=0.75]
    {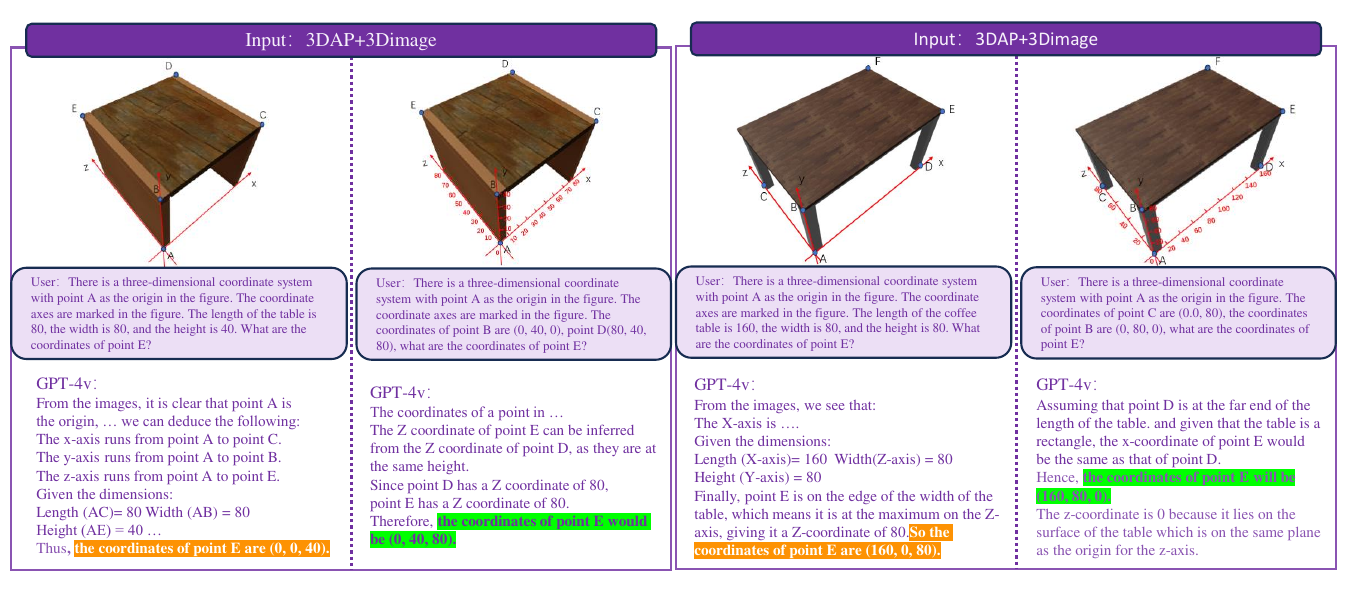}
   \caption{The ablation experiments under two conditions of input image annotation:(left)adding only the coordinate system without any scale markings. (right)includes labels for both the coordinate system and the scale.}
   \label{fig:ab1}
\end{figure*}

\textbf{\textit{2D to 3D Point Matching}}
is a sophisticated visual positioning technique, meticulously crafted to compare and align features of two-dimensional images with three-dimensional models. The core of this task is to find the matching feature points between the 2D image and the 3D model and use these matching points to estimate the precise position and orientation of the 2D image relative to the 3D model.\cite{song2021recalling,li2023geo,ma2021image,nadeem2020unconstrained} This process not only involves matching points on a two-dimensional image to their corresponding points in a three-dimensional model or scene but also requires taking into account complex changes in perspective and scale\cite{feng20192d3d,wang2021p2}.

In terms of viewing angle, there may be significant differences in viewing angle between 2D images and their corresponding 3D models or scenes. This means that even the same object may show different forms and details in different viewing angles, which may include partial occlusion and distortion of viewing angles. This change in perspective requires the algorithm to be able to recognize and adapt to different observation conditions to ensure accurate feature matching in various situations. In terms of scale, due to the difference in distance between the 2D image and the object during the shooting process, the resulting scale difference is also an important consideration. This difference may lead to different sizes of the same object in the image, affecting the recognition and matching of features. Scale variation may lead to loss or deformation of details in the image, which needs to be overcome by multi-scale analysis \cite{zhang2022point}and scale-invariant feature extraction methods\cite{lindeberg2012scale,lim20193d,DBLP:journals/corr/abs-2112-10258}. 

In order to effectively solve the problem of perspective and scale differences between 2D images and 3D models, a deep understanding of the relationship between 2D and 3D data is required, as well as a high degree of adaptability to perspective and scale differences. Through the use of advanced technologies and algorithms, precise visual positioning can be achieved, thus playing an important role in a variety of practical applications.


\begin{table*}[htbp]
	\centering
	\caption{A comparative analysis was conducted on the task of 20 to 3D point reconstruction.}
	\label{tab:1}  
	\begin{tabular}{ccccccc}	\hline\noalign{\smallskip}	 & chair& table & sofa & cabinet & overall
  \\
\noalign{\smallskip}\hline\noalign{\smallskip}
GPT-4v&0.25&0.28&0.33 &0.29&0.29
 \\
 GPT-4v+3DAP & 0.83 & 0.79 & 0.92&0.86 &0.85\\\noalign{\smallskip}\hline
	\end{tabular}
\end{table*}

\begin{table*}[htbp]
	\centering
	\caption{An ablation experiment in which a scale mark is added to the input image}
	\label{tab:2}  
	\begin{tabular}{cccc ccc}	
 \hline\noalign{\smallskip}	 &chair&table&sofa&cabinet&overall
  \\
\noalign{\smallskip}\hline\noalign{\smallskip}
GPT-4v+3DAP-scale&0.71&0.67&0.75 &0.63&0.64
 \\
 GPT-4v+3DAP & 0.86 & 0.78 & 0.83&0.88 &0.83 \\\noalign{\smallskip}\hline
	\end{tabular}
\end{table*}

\textbf{\textit{3D Object Detection}}
is a pivotal task in the field of computer vision. It involves the technology of recognizing and pinpointing objects within a three-dimensional space. Going beyond the capabilities of traditional 2D object detection, 3D object detection is adept at not only identifying the type and location of objects but also accurately gauging their size, orientation, and depth in a three-dimensional context.\cite{arnold2019survey,liang2021survey,ahmed2020density,wang2020overview} This field is highly dynamic and rapidly evolving, finding extensive applications in areas such as autonomous driving, robotic navigation, augmented reality, and virtual reality.

In the realm of autonomous driving, 3D object detection plays a critical role in identifying other vehicles, pedestrians, and obstacles on the road, thereby significantly enhancing driving safety.\cite{arnold2019survey,qian20223d,li20236dof,wang2023multi} Within the domain of robotic navigation, 3D object detection aids robots in comprehending and adapting to complex environments\cite{DBLP:journals/corr/GregorioS17,chaplot2020object}, facilitating effective obstacle avoidance and path planning. Furthermore, in AR and VR, 3D object detection considerably enriches user interaction with digitally created content, making these experiences more immersive and intuitive.\cite{choudhary2020multi,ahmadyan2020instant,xu20233difftection,ghasemi2022deep} As such, the advancements in 3D object detection are not only technologically impressive but also pivotal in shaping the future of intelligent systems across various sectors.

\subsection{Comparison of 3DAP with  state of the art}
This research undertook detailed experiments within the three principal 3D task domains previously outlined. This section will meticulously detail the experimental procedures employed in each of these task scenarios.

\textbf{\textit{2D to 3D Point Reconstruction}}

Figure \ref{fig:ex1} presents a detailed experimental comparison between the input marks only the key points and the image annotated using the 3DAP method in the task of 2D to 3D Point Reconstruction. When the input image solely highlights the location information of key points, GPT-4v's capability to perceive the spatial dimensions of the 3D image is somewhat limited. In this scenario, it is recommended for users to provide additional modeling cues to enhance the representation of 3D spatial information. When images annotated with the 3DAP method are used as input, there is a notable enhancement in GPT-4v's ability to interpret three-dimensional space. This leads to a more accurate reconstruction of the direction and coordinate positions of other points in the 3D space, demonstrating the effectiveness of the 3DAP method in enriching GPT-4v's spatial understanding and reconstruction accuracy.

\textbf{\textit{2D to 3D Point Matching}}

Figure \ref{fig:ex2} shows the experimental comparison between the input original image and the input image marked by 3DAP method in 2D to 3D point matching task. When multiple images are fed into the system without any preprocessing, the GPT-4v model demonstrates insufficient accuracy in discerning the precise positions and orientations of points. However, with the inclusion of both 3D and 2D images marked by the 3DAP method, the GPT-4v shows enhanced capability in spatial modeling. It leverages the 3D coordinates provided by 3DAP to accurately pinpoint key positions and directions. Moreover, it effectively identifies the corresponding key points and their spatial interrelations within the 2D image.

\textbf{\textit{3D Object Detection}}

Figure \ref{fig:ex3} presents a comparative analysis between the input original image and the input image marked by 3DAP method in 3D Object Detection task. In the case that only the original unlabeled scene image is entered and the text description contains only reference coordinate information, the GPT-4v prompts the user to add more 3D modeling details, such as axis and scale size information. However, when the input is a scene image labeled by the 3DAP method, GPT-4v is able to more accurately understand the spatial proportions in the image and give a more accurate judgment of the coordinate range of the target object.

\subsection{Quantitative Results}
We conducted experiments on 2D to 3D Point Reconstruction using the 3DAP-based data set we constructed,  and the results are shown in Table \ref{tab:1}. Two types of inputs were used: the original image and an image annotated with the 3DAP mark. For each, key dimensional details of the object, including height, width, and length, were described. The results showed that the quality of reconstruction from the 3DAP-prompted images was significantly superior to that from the original images, highlighting the effectiveness of the 3DAP prompting in enhancing 3D reconstruction accuracy. It shows the importance and effectiveness of 3DAP prompt method for GPT-4v's ability to unleash 3D space.

\section{Ablation Study}
\label{sec:ablation}

In this paper, we conducted ablation experiments for 2D to 3D Point Reconstruction task and confirmed the significance of simultaneous marking of coordinate system and scale in 3DAP. Figure \ref{fig:ab1} illustrates the ablation experiments conducted on the input image, demonstrating two scenarios: one with the addition of a coordinate system without scale markings, and the other with both a coordinate system and scale markings labeled, ablation experiments were conducted on the small data set 3DAP-Data proposed by us. The findings are presented in Table \ref{tab:2}. It was observed that the integration of scale indicators enhances the precision of GPT-4v in assessing point-specific details, leading to an advancement in the model's capability to ascertain three-dimensional spatial positioning.

The experimental results show that GPT-4v can accurately describe the corresponding coordinate information when only the coordinate axis is labeled and the quantitative values of the entity "height and width" are added to the description language of the image, but there is a deviation in the analysis of the key point coordinates. When the coordinate system and scale are marked in the image at the same time, and the point coordinate information is accurately provided in the description language, GPT-4v can accurately analyze the three-dimensional spatial position relationship and key point coordinates expressed by the entity.

\section{Conclusion}
\label{sec:conclusion}
We propose a new visual prompt method called 3DAP, which is applied to GPT-4v. By annotating the 3D coordinate system and scale information of 3D images, GPT-4v can more effectively align 3D images with their corresponding coordinate systems and grasp the positional relationships among various points within these 3D images to unleash the 3D Spatial Task Capabilities of GPT-4V. We conducted experiments and observations on the proposed visual prompt method to demonstrate that 3DAP can maximize the potential of GPT-4v's 3D understanding capabilities. In our future work, we aim to extend the application of 3DAP to other large language models, enabling them to comprehend the rich and tangible world of three-dimensional perception.
{
    \small
    \bibliographystyle{ieeenat_fullname}
    \bibliography{main}
}

\clearpage
\section *{Appendix}
\label{sec:append}
\appendix
In this supplementary material, we provide a detailed overview of our data set and demonstrate our performance across three key tasks: conversion from 2D to 3D reconstruction, matching 2D to 3D points, and detecting 3D objects utilizing the 3DAP prompt methodology. We further substantiate the efficacy of the 3DAP approach with more examples.

\section{Datasets}
\label{sec:datasets}

\begin{figure}[H]
  \centering
    \includegraphics[scale=0.5]
    {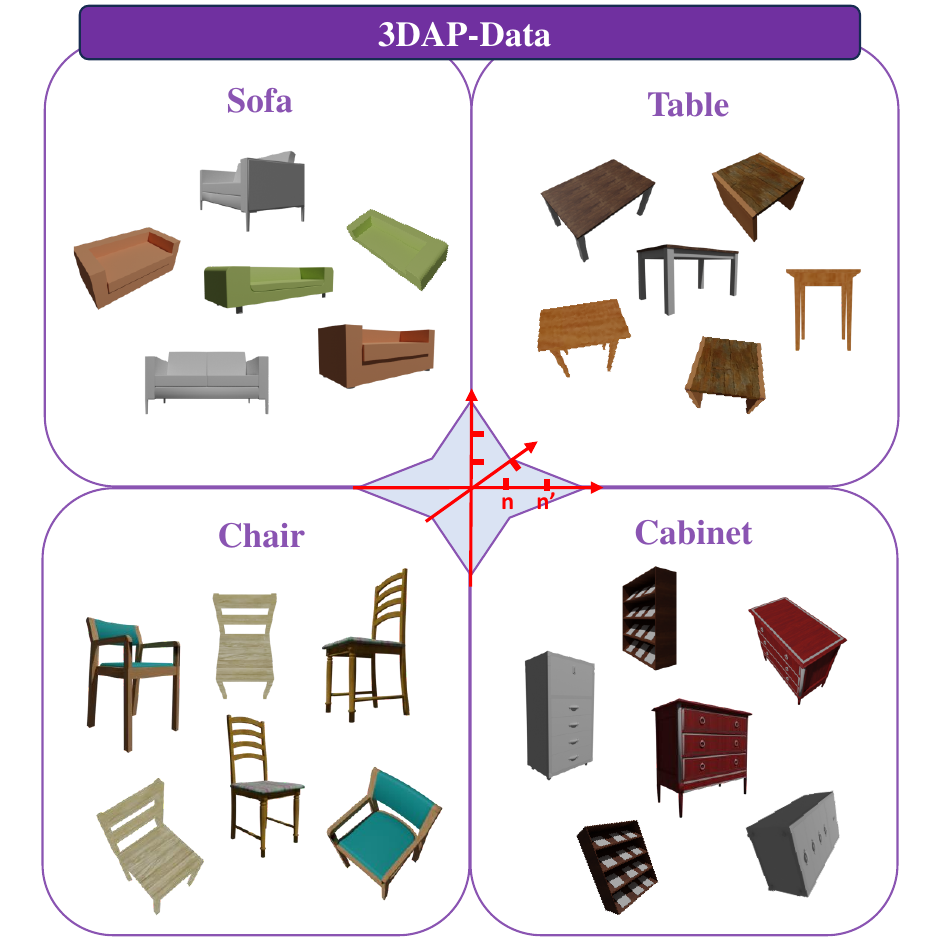}
   \caption{3DAP-Data contains four categories, each containing multiple entities, each of which has multiple view representations. The 3D coordinate system and scale information of multi-view 3D images are manually marked.}
   \label{fig:dataset}
\end{figure}

We have developed a compact dataset, named 3DAP-Data, derived from the 3DAP visual prompt methodology. As is shown in Figure  \ref{fig:dataset} , this dataset encompasses a selection of objects commonly encountered in everyday living environments, such as chairs, tables, cabinets, and sofas. These objects were specifically selected for their distinct vertical lines and plane relationships, offering a rich tapestry of geometric insights.

A key feature of 3DAP-Data is its diversity and breadth. Each category of objects within the dataset is represented through a wide array of instances, each differing in shape and color.  This variety ensures a comprehensive understanding of how these common objects can vary in appearance and structure. Furthermore, each instance within the dataset is represented through multiple views. This multi-view approach enables a more holistic understanding of the objects, capturing them from various angles and perspectives.
We manually labeled the three-dimensional coordinates and scale information of multi-view 3D images, enabling the annotation of spatial dimension information of entities in images, the dataset serves not just as a collection of images, but as a rich source of spatial information. It significantly enhances GPT-4's ability to interpret and comprehend spatial information.

\section{Use Case}
\label{sec:case}
\begin{figure*}[htbp]
  \centering
    \includegraphics[scale=0.62]
    {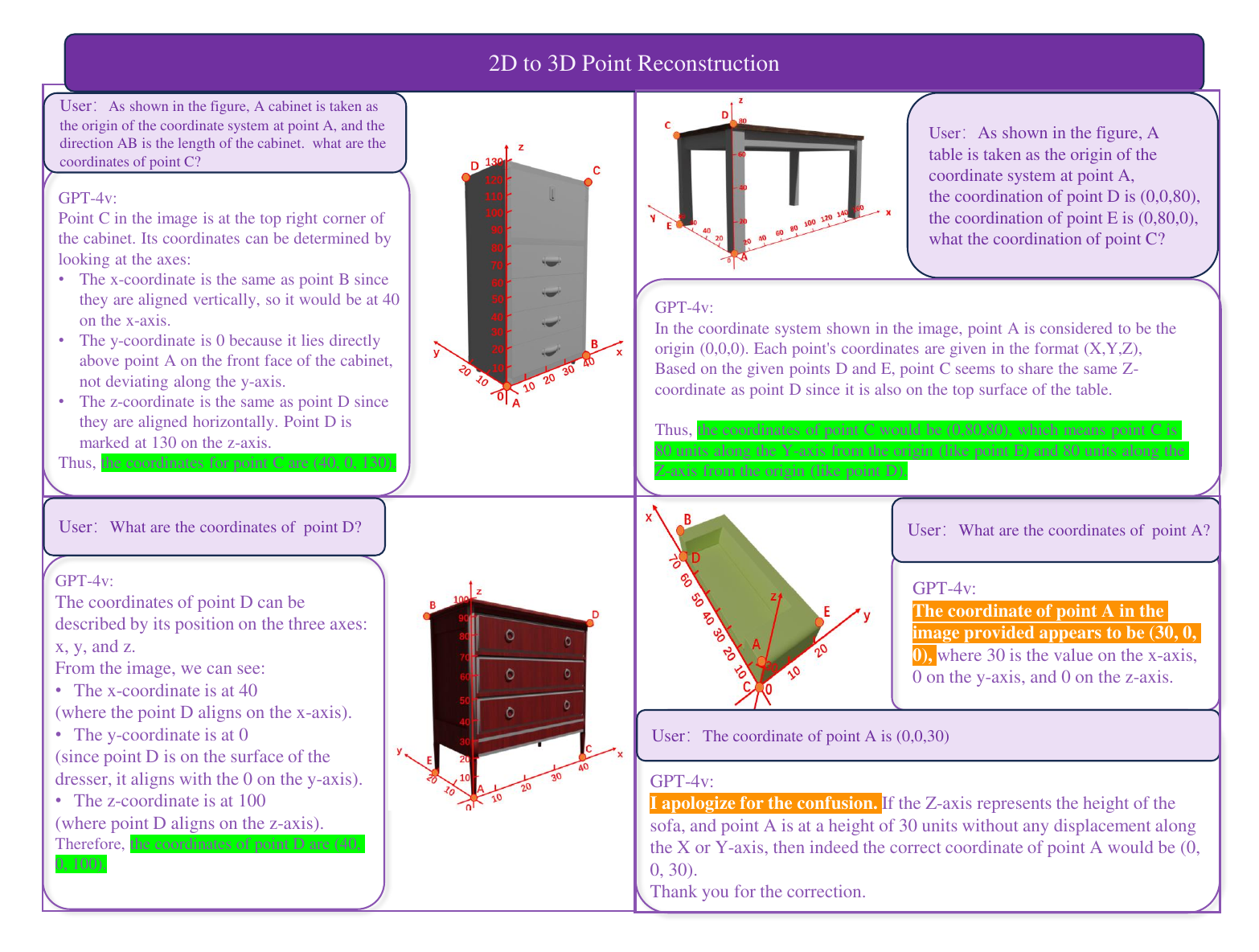}
   \caption{Effect on 2D to 3D Point Reconstruction task}
   \label{fig:ap1}
\end{figure*}

In this section, we present the effects of using the 3DAP method on each of the three tasks.
\paragraph{2D to 3D Point Reconstruction}
Figure \ref{fig:ap1} illustrates the effect on 2D to 3D Point Reconstruction task. The process involves inputting an image prompting with 3DAP and key points, modifying the input language description, and then observing GPT-4v's response. 

In the case at the upper left corner, we select the side view of a cabinet. The description emphasizes the origin position and axis direction of the three-dimensional coordinate system within the object. GPT-4v successfully deciphers the point coordinate information, analyzes the 3D spatial data of the physical object, and accomplishes the reconstruction of the 3D point location.

In the case at the upper right corner,  we choose the side elevation view of a table. The description includes the origin and coordinates of certain key points. GPT-4v evaluates the axis direction and precisely identifies the coordinates of key points not initially provided.

In the case in the lower left corner, we focus on the side view of the cabinet. However, the description directly queries the coordinates of key points. GPT-4v leverages the 3DAP prompting on the input image to ascertain the three-dimensional spatial positioning of the object and accurately determines the key point coordinates.

Finally, in the case in the lower right corner, we select the top view of a sofa. In this instance, the height dimension of the sofa aligns with the viewing angle, making it challenging to discern. GPT-4v initially misinterprets the height dimension, but after manual correction, it adjusts its analysis accordingly.

\begin{figure*}[htbp]
  \centering
    \includegraphics[scale=0.34]
    {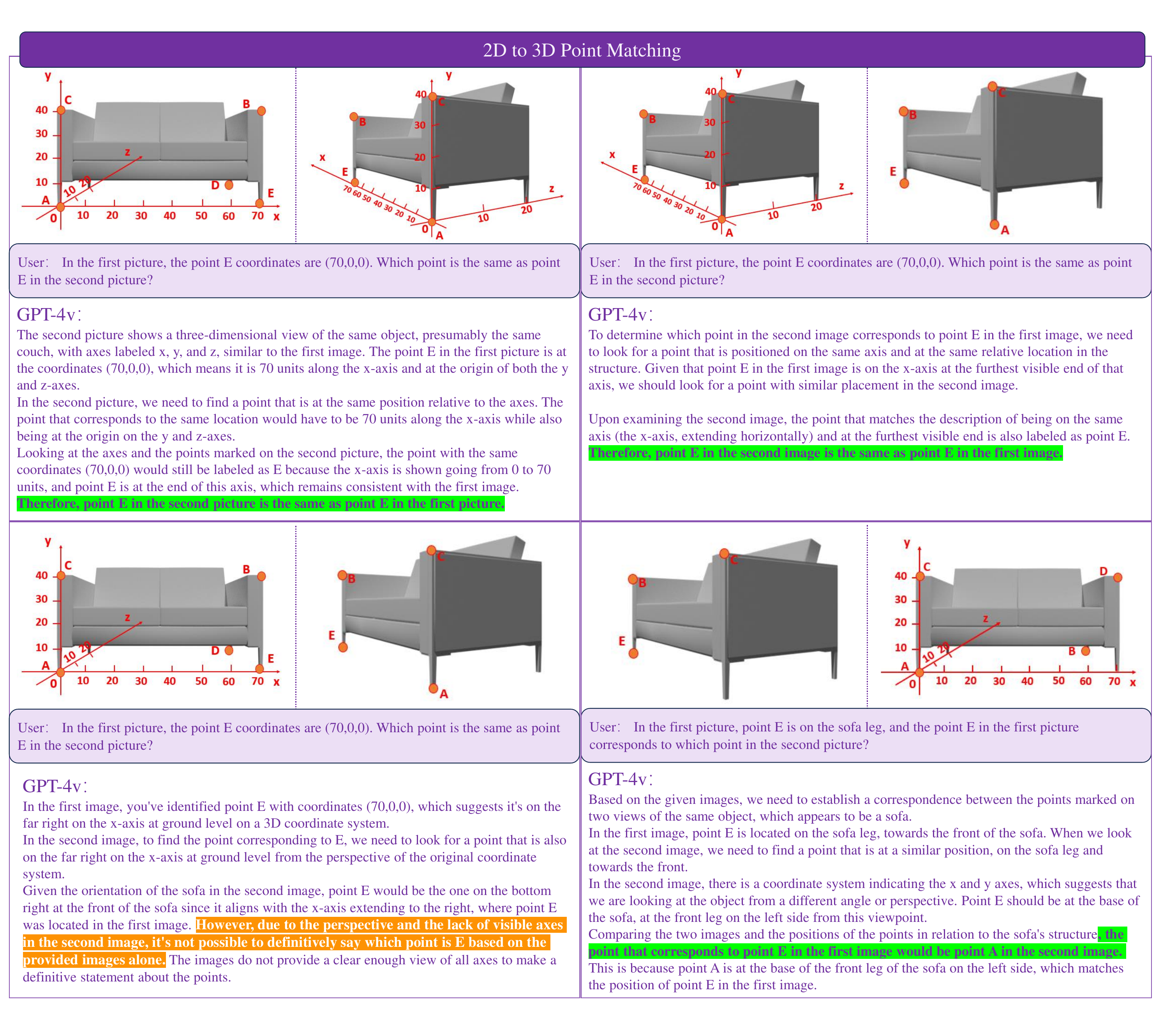}
   \caption{Effect on 2D to 3D Point Matching task}
   \label{fig:ap2}
\end{figure*}

\paragraph{2D to 3D Point Matching}
Figure \ref{fig:ap2} illustrates the effect on 2D to 3D Point Matching task.  In the provided input, the image and its key point data are presented through the utilization of the 3DAP marking technique. This input image encompasses both identical and varying view angles of the same entity, modifying the input language description and image, and then observing GPT-4v's response.

In the case at the upper left corner, we select the main view of a sofa entity and the side view of a sofa entity. The key points are marked in both images, and the spatial information is marked with 3DAP prompt method. At the same time, the coordinate information of key points is given in the language description. GPT-4v can accurately analyze the spatial positioning of the first image and find the key points corresponding to the first image in the second image.

In the case at the upper right corner, we select the same side view of the sofa entity. The figure on the left uses 3DAP prompt method to mark spatial information and key points, while the figure on the right only marks key points and gives the coordinate information of key points in the language description. GPT-4v finds the key points corresponding to the first image in the second image by analyzing the axis orientation and position of the base image on the left.

In the case at the lower left corner, we select the main view of a sofa entity and the side view of a sofa entity. The main view uses 3DAP prompt method to mark spatial information and key points, while the side view only marks key points. The coordinate information of the key points is given in the language description, GPT-4v analyzes the axis direction and position of the left reference picture, and in the process of finding the key points matching the left picture from the right picture, GPT-4v prompts the user to provide more annotation information in the right picture, such as the axis.

Finally, in the case in the lower right corner, we select the side view of a sofa entity and the main view of a sofa entity. The left-side view is used as a reference figure, and only key points are marked. the main view on the right uses 3DAP to mark spatial information and key points, and gives the description of key points in the language description, such as "point E is on the sofa leg". GPT-4v analyzes the spatial position of key points in the benchmark image on the left. According to the coordinate axis and point information marked by the image on the right, find the key points corresponding to the benchmark image on the left.

\begin{figure*}
  \centering
    \includegraphics[scale=0.48]
    {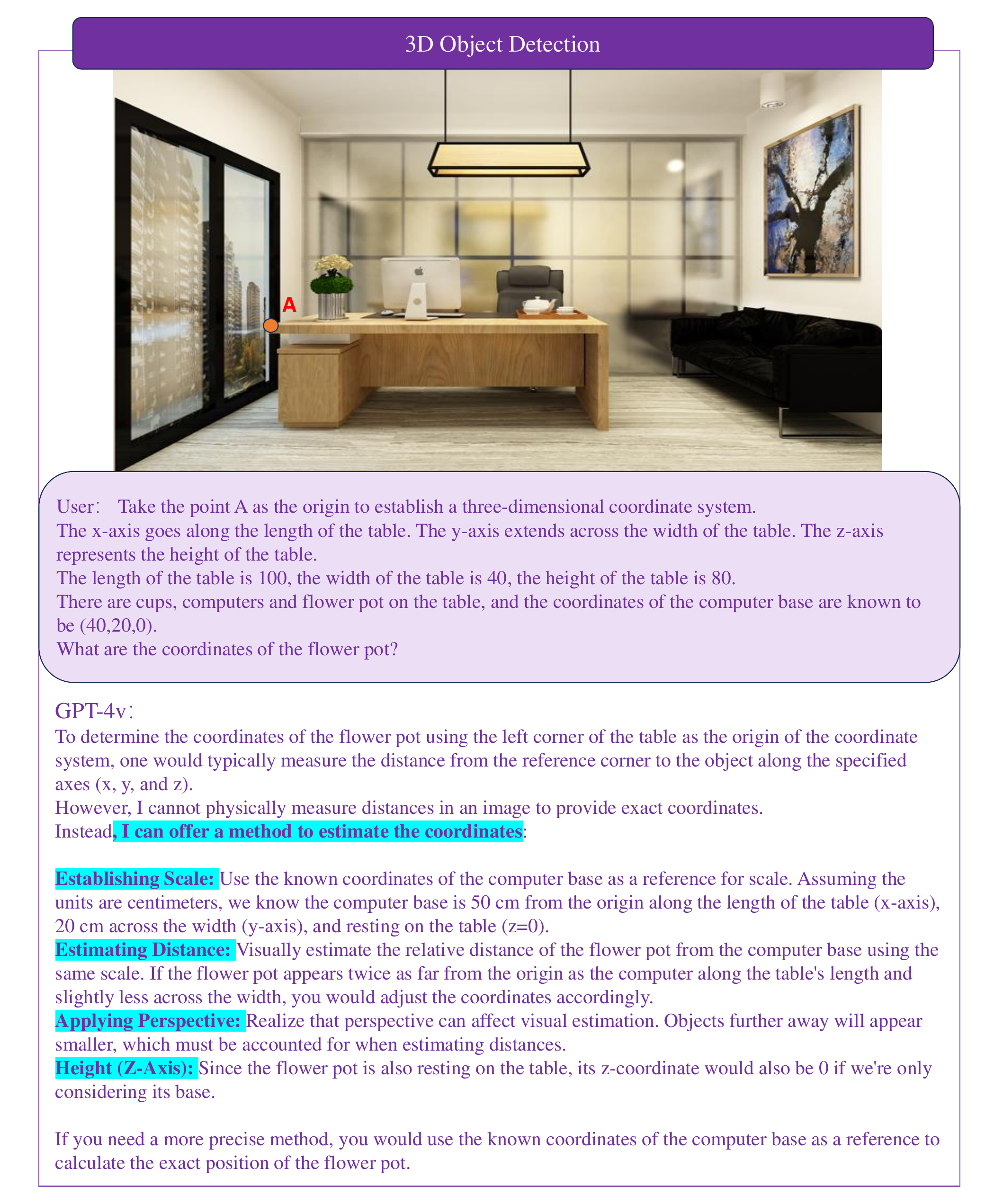}
   \caption{Effect on 3D Object Detection task without 3DAP}
   \label{fig:ap3}
\end{figure*}

\begin{figure*}
  \centering
    \includegraphics[scale=0.34]
    {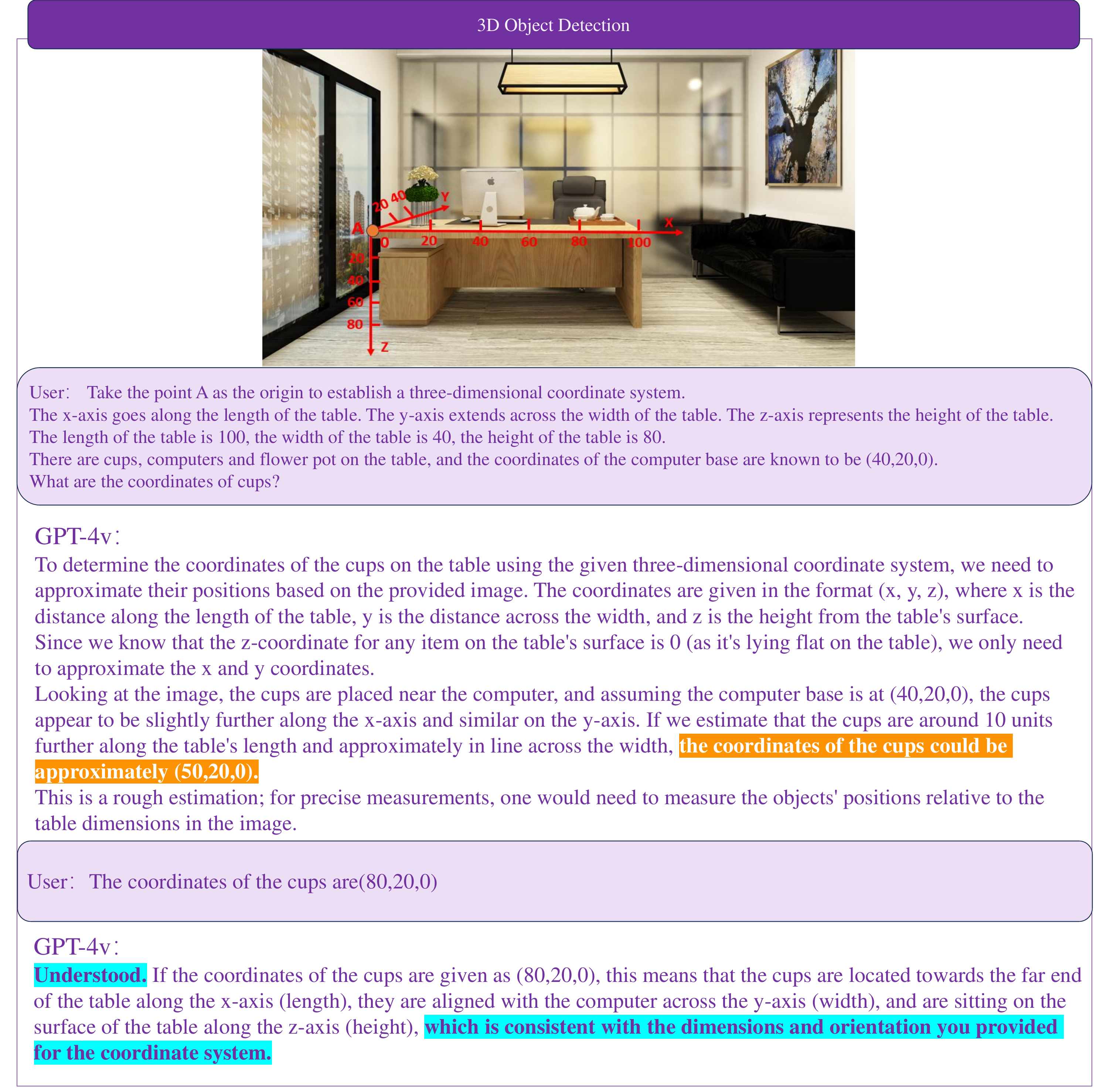}
   \caption{Effect on 3D Object Detection task with 3DAP}
   \label{fig:ap4}
\end{figure*}

\paragraph{3D Object Detection}
Figure \ref{fig:ap3} and Figure \ref{fig:ap4} illustrate the effect on 3D Object Detection task. In Figure \ref{fig:ap3}, we input an image depicting an indoor scene. This image is uniquely marked with the origin of the coordinate axis and provides a brief description of the scene construction in the text description, including the direction of the coordinate axis construction and the length, width, and height measurements of the table in the scene. Utilizing these elements, GPT-4v endeavors to decipher the spatial relationships within the scene. Furthermore, it provides a detailed methodology and sequential steps for establishing a coordinate system. It encourages users to input more precise information, thereby enhancing GPT-4v's understanding of the spatial dimensions depicted.

Figure \ref{fig:ap4}, we input an indoor scene picture prompted with 3DAP, and explain in the text description the direction of the construction of the coordinate axis and the length, width, and height of the table in the scene as shown in Figure \ref{fig:ap3}. GPT-4v can accurately analyze the direction of the coordinate axis and the position information of the object in space. There is some deviation in the recognition of detailed and precise scales, but with human correction, GPT-4v is capable of analyzing and agreeing with the correction results.

\end{document}